\documentclass[manuscript, screen, sigconf, nonacm]{acmart} 
\usepackage{subcaption}

\begin{document}

\title{Exploring Dynamic Novel View Synthesis Technologies for Cinematography}

\author{Adrian Azzarelli}
\email{a.azzarelli@bristol.ac.uk}
\orcid{1234-5678-9012}
\affiliation{%
  \institution{University of Bristol}
  \city{Bristol}
  \country{UK}
}

\author{Nantheera Anantrasirichai}
\email{n.anantrasirichai@bristol.ac.uk}
\orcid{1234-5678-9012}
\affiliation{%
  \institution{University of Bristol}
  \city{Bristol}
  \country{UK}
}

\author{David R Bull}
\email{david.bull@bristol.ac.uk}
\orcid{1234-5678-9012}
\affiliation{%
  \institution{University of Bristol}
  \city{Bristol}
  \country{UK}
}

\renewcommand{\shortauthors}{Azzarelli et al.}

\begin{abstract}
  Novel view synthesis (NVS) has shown significant promise for applications in cinematographic production, particularly through the exploitation of Neural Radiance Fields (NeRF) and Gaussian Splatting (GS). These methods model real 3D scenes, enabling the creation of new shots that are challenging to capture in the real world due to set topology or expensive equipment requirement. This innovation also offers cinematographic advantages such as smooth camera movements, virtual re-shoots, slow-motion effects, etc. This paper explores dynamic NVS with the aim of  facilitating the model selection process. We showcase its potential through a short montage filmed using various NVS models.
\end{abstract}

\maketitle

\section{Introduction}
\begin{figure*}[ht]
    \centering
    \includegraphics[width=\linewidth]{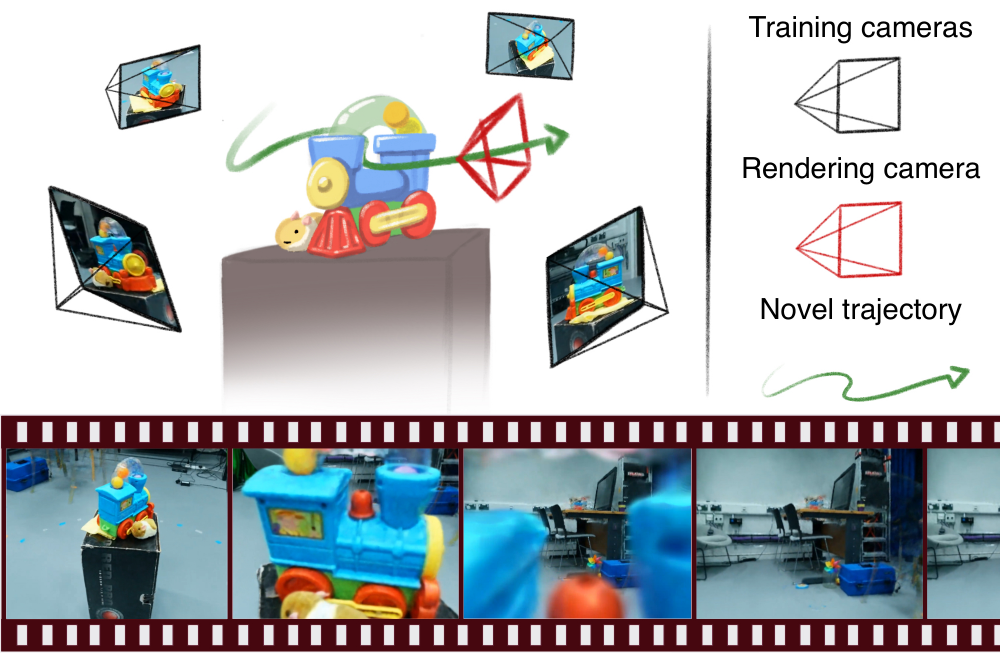}
    \Description{The main figure is an 3-D vector illustration of a toy train surrounded by images used for training. A synthesized camera path and a novel rendering camera of a fly-through shot are shown in green and red color, respectively. The bottom subfigure shows several frames of the rendered fly-through shot.}
    \caption{Fly-through shot on a toy-scene}
    \label{fig: fly-through}
\end{figure*}

Novel view synthesis (NVS) creates new images and videos of a scene from viewpoints that are not captured in the original data by utilizing a virtual representation of a real 3D scene. The recent success of learning-based techniques like NeRF and GS, has enabled the production of high-quality volumetric scenes from existing image and video data. This offers significant advantages in cinematography: (1) enabling smooth camera movement without requiring gimbals or dollies; (2) supporting virtual re-shoots, thus eliminating the need for expensive in-person re-shoots; (3) producing slow-motion effects without requiring special equipment; (4) altering scene geometry and aesthetics without manual intervention, reducing the burden on visual effects and compositing artists \cite{Yuan_Sun_Lai_Ma_Jia_Gao_2022,Wang_Chai_He_Chen_Liao_2022,Sun_Wang_Zhang_Li_Zhang_Liu_Wang_2022,Song_Choi_Do_Lee_Kim_2023,Kuang_Luan_Bi_Shu_Wetzstein_Sunkavalli_2023,Bao_Zhang_Yang_Fan_Yang_Bao_Zhang_Cui_2023,Chen_Lyu_Wang_2023,He_Yuan_Zhu_Dong_Bo_Huang_2024}; and (5) introducing/correcting camera-based artefacts like de-blurring \cite{Wang_Zhao_Ma_Liu_2023} and lens distortions \cite{Kim_Gu_Choo_2024, Xian_Božič_Snavely_Lassner_2023,Moreau_Piasco_Tsishkou_Stanciulescu_deLaFortelle_2022}. More creative uses have also been explored, like cinematic style transfer \cite{Wang_Courant_Shi_Marchand_Christie_2023}, the discovery of new cinematographic shots \cite{Skartados_Yucel_Manganelli_Drosou_SaGarriga_2024} and modeling live sport for action-replay \cite{Lewin_Vandegar_Hoyoux_Barnich_Louppe_2023}. In Figure \ref{fig: fly-through}, we demonstrate this by training a static NeRF toy scene using only 36 images to synthesize a fly-through shot that would otherwise require special equipment.

Considering the primary goal in cinematography is to capture scenes or assets in motion, our focus is on dynamic NVS research. We explore these new technologies and evaluate their meaning and effectiveness in relation to cinematographic content acquisition. We specifically explore:
\begin{enumerate}
    \item Dynamic Representations
    \item Dynamic scenes vs Animate-able human assets
    \item Data Acquisition
\end{enumerate}
Subsequently, we film a short montage captured using various static and dynamic NVS models to showcase their potential in cinematographic production.

\section{Technical Background}
\begin{figure*}[t]
    \centering
    \includegraphics[width=\linewidth]{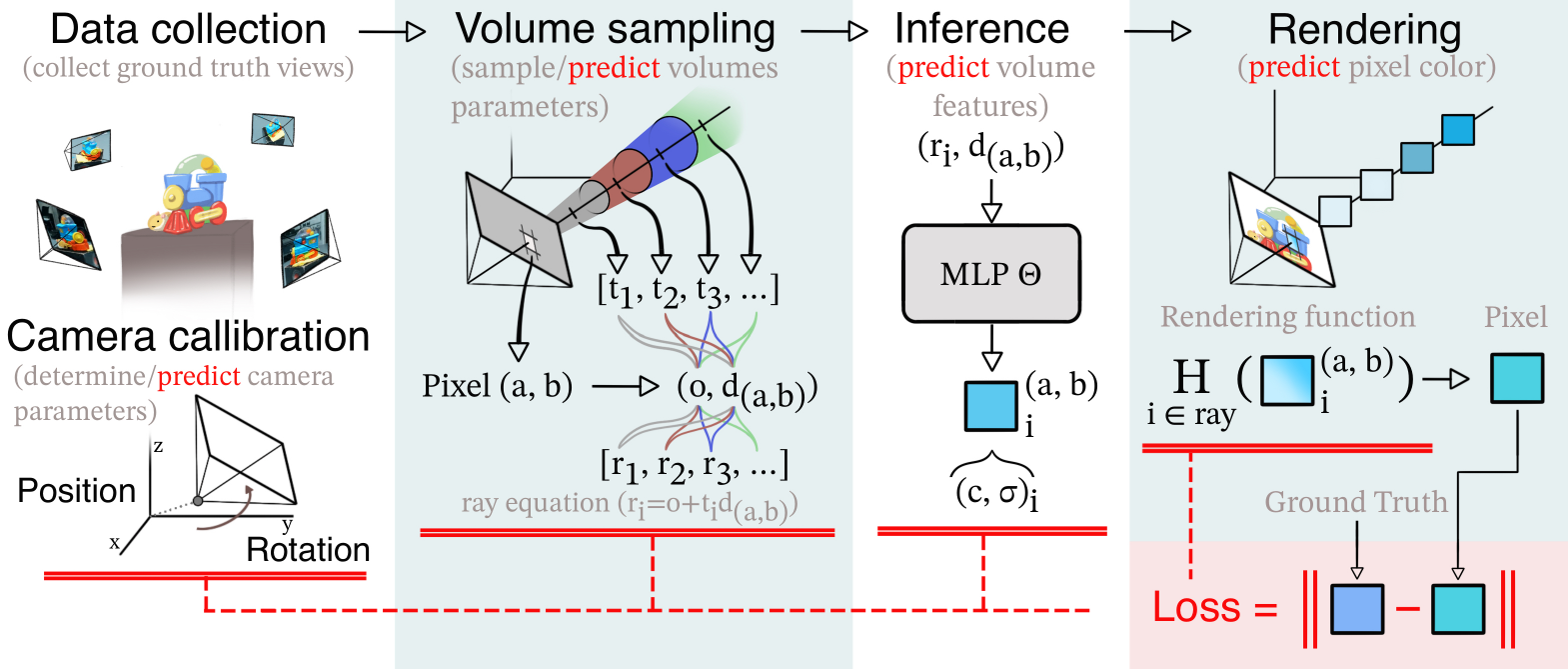}
    \Description{The image contains four columns. The first column illustrates the data collection and camera calibration process, where images of a real set of collected and the position and orientation of each image is determined. The second column illustrates the volume sampling process where volume segments along a 3-D ray are sampled leading to determining the origin and viewing direction of each sample. The third column illustrates the inference step where origin and direction are passed into an MLP and a color and density value is estimated. The fourth column consists of rendering the ray-colors for each pixel-ray and propagating the loss (comparing the predicted pixel to a ground truth value) to the camera calibration, volume sampling, inference and rendering steps.}
    \caption{An illustration of the classical NVS pipeline. The learnable parts of the pipeline are highlighted in red}
    \label{fig: pipeline}
\end{figure*}

Figure \ref{fig: pipeline} illustrates the main steps involved in generating a neural 3D representation to achieve NVS. A collected image dataset first requires calibration to determine the position and rotation of each image with respect to the world coordinate space. Then, for each view in the training set we sample the in-frustum volumetric space and pass the position and viewing direction of each volume to a neural net that estimates the color and density of each volume-sample. Finally, we render the in-frustum volumes with respect to depth and backpropagate a loss by evaluating the predicted images against the ground truth images.

\begin{figure*}[t]
    \centering
    \begin{subfigure}{\linewidth}
        \includegraphics[width=\linewidth]{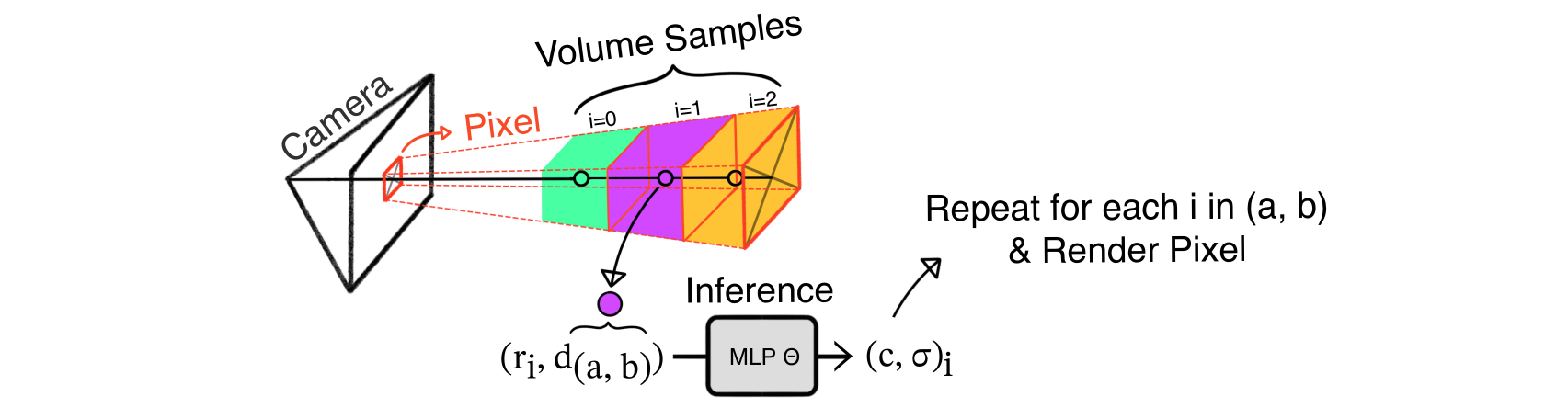}
        \caption{\textbf{Neural Radiance Fields} sample in-frustum volumes to estimate a color and density value for each sample projected along a pixel-ray}
    \end{subfigure}
    \begin{subfigure}{\linewidth}
        \includegraphics[width=\linewidth]{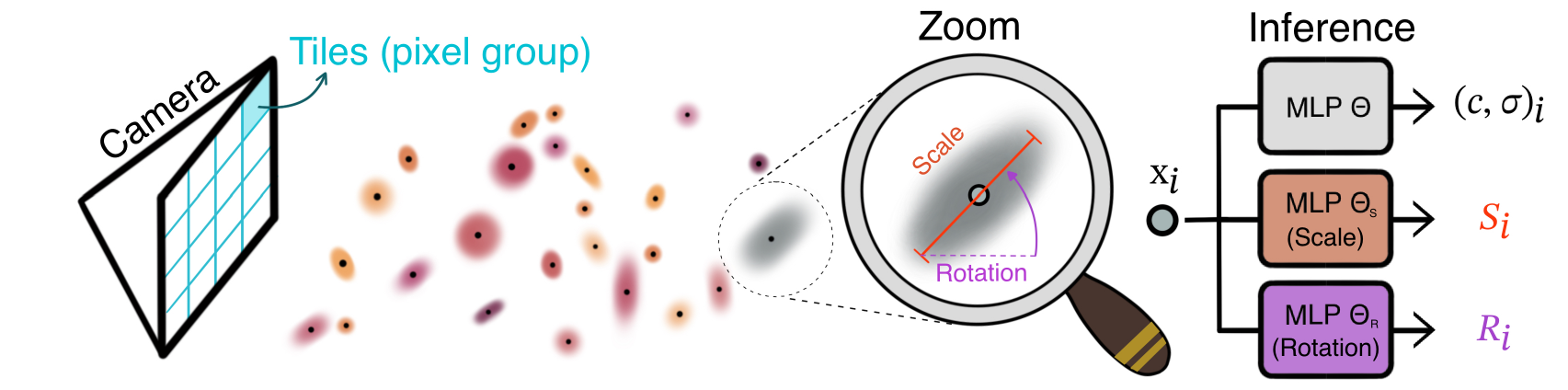}
        \caption{\textbf{Gaussian Splatting} uses a point cloud to estimate the color, density and covariance matrix (scale and rotation) of each 3D Gaussian point, which are rendered via tile-splatting}
    \end{subfigure}
    \Description{Two subfigures are shown. The first subfigure illustrates the NeRF approach, where a ray is projected from a pixel belonging to a specific camera/view and the global position of the volume segments along the ray are used to estimate the color of the pixel. The second subfigure illustrates the GS approach, where a distribution of gaussian points appear in front of a camera view and for each point the color, density, scale and rotation are predicted.}
    \caption{An illustration of the original NeRF and GS representations}
    \label{fig: nerf vs gaussiansplatting}
\end{figure*}

The predominant challenge is how best to represent the volumetric space and achieve view-dependent effects, e.g. reflections from non-Lambertian materials. Addressing this, the two contributions that have shaped the NVS landscape are NeRF and GS, which are discussed below and accompanied by Figure \ref{fig: nerf vs gaussiansplatting}.

\subsection{Neural Radiance Fields} \citet{mildenhall2021nerf} propose the NeRF representation using an implicit coordinate multilayer perceptron (MLP) \begin{math}\Theta(\cdot)\end{math}, defined as follows.
    \begin{equation*}
        \Theta(\mathbf{r_i}, \mathbf{d}_{a,b}) \rightarrow (\mathbf{c}_i, \sigma_i),
    \end{equation*} takes the position \begin{math}\mathbf{r}_i = [x,y,z]_i\end{math} along a ray \begin{math}\mathbf{r}\end{math}, and viewing direction \begin{math}\mathbf{d}_{a,b}\end{math} of a volume \begin{math}i\end{math} to predict the color \begin{math}\mathbf{c}_i\end{math} and density \begin{math}\sigma_i\end{math}. As each ray corresponds to a pixel \begin{math}(a,b)\end{math} in an image \begin{math}\mathbf{H}(\mathbf{r})\end{math} with a focal point at \begin{math}\mathbf{o}\end{math}, the relationship between \begin{math}\mathbf{r_i}\end{math} and \begin{math}\mathbf{d}_{a,b}\end{math} can be written as
    \begin{equation*}
        \mathbf{r}_i= \mathbf{o} +t_i \mathbf{d}_{a,b}.
    \end{equation*}
    This enables learning view-dependent effects. Consequently, an image can be rendered with
    \begin{equation}\label{eq: nerf}
        \hat{\mathbf{H}}(\mathbf{r}) = \sum^{t_{far}}_{i={t_{near}}} (1 - \exp(-\sigma_i \delta_i)) \mathbf{c_i} \exp({\sum^{i-1}_{j=t_{near}} - \sigma_j \delta_j}),
    \end{equation}
    where \begin{math}\delta_i\end{math} is the width of each ray-volume sample.

\subsection{Gaussian Splatting} \citet{kerbl20233d} propose the GS representation, where a differentiable point cloud is used to model the position of 3D Gaussian volumes, \begin{math}\mathbf{x}_i \in \mathbf{x}\end{math}, in 3D space using,
    \begin{equation}
        G(\mathbf{x}) = exp(- \frac{1}{2} (\mathbf{x})^T \Sigma ^{-1} (\mathbf{x})),
    \end{equation}
    where 
    \begin{equation}
        \Sigma' = JW \Sigma_i W^T J^T
    \end{equation}
    is a \begin{math}3\times3\end{math} covariance matrix dependent on the viewing transform \begin{math}W\end{math} and the Jacobian of the affine projection transform \begin{math}J\end{math}. In GS, \begin{math}\Sigma'\end{math} is simplified into a \begin{math}2 \times 2\end{math} matrix by ignoring the third row and column.
    
    For each point \begin{math}\mathbf{x}_i\end{math}, a rotation \begin{math}R_i\end{math} and scale \begin{math}S_i\end{math} are approximated with an MLP allowing for \begin{math}\Sigma_i = R_i S_i S_i^T R_i^T\end{math}. The color \begin{math}\mathbf{c}_i\end{math} and density \begin{math}\sigma_i\end{math} of each sample are determined using a separate MLP. This forms the basis of the GS representation, which is rendered with a differentiable tile-splatting strategy where an image, with properties \begin{math}W\end{math} and \begin{math}J\end{math}, is split into \begin{math}16 \times 16\end{math} tiles, the Gaussians \begin{math}G(\mathbf{x})\end{math} are ordered with respect to depth and finally rendered using the equation:
    \begin{equation}
        \hat{\mathbf{H}}(\mathbf{x}) = \sum_{i\in N} \mathbf{c}_i \alpha_i \prod^{i-1}_{j=1}(1-\alpha_j) ,
    \end{equation}
    where \begin{math}\alpha_i\end{math} is given by multiplying \begin{math}\Sigma_i\end{math} with the learned point-opacity \begin{math}\sigma_i\end{math}.

\subsection{NeRF vs. GS}
Comparing the two representations and rendering methods, GS is significantly faster and shows improved quality over NeRF \cite{kerbl20233d}. However, solutions have been proposed that address the slow NeRF inference problem \cite{Gu_Jiang_Li_Lu_Zhu_Shah_Zhang_Bennamoun_2024,Kurz_Neff_Lv_Zollhöfer_Steinberger_2022,Yu_Li_Tancik_Li_Ng_Kanazawa_2021,Wang_Hu_He_Wang_Yu_Tuytelaars_Xu_Wu_2023}. Despite these advances, GS struggles to produce acceptable depth estimates without proper initialization \cite{foroutan2024does}, which means that its practical use may be limited by the need for ground truth depth data. This can be acquired with additional depth sensors or by relying on neural depth estimators \cite{Fan_Cong_Wen_Wang_Zhang_Ding_Xu_Ivanovic_Pavone_Pavlakos_etal_2024}. 

\section{Dynamic NVS for Cinematography}

\subsection{Dynamic Representations}
With the vast collection of existing research on how to best represent a dynamic 3D scene, choosing the ideal representation for a specific cinematic application is a challenging task. This is compounded by the need to concurrently learn view-dependent and temporal effects, which are difficult to disentangle when relying solely on images for supervision.

\begin{figure*}[t]
    \centering
    \includegraphics[width=\linewidth]{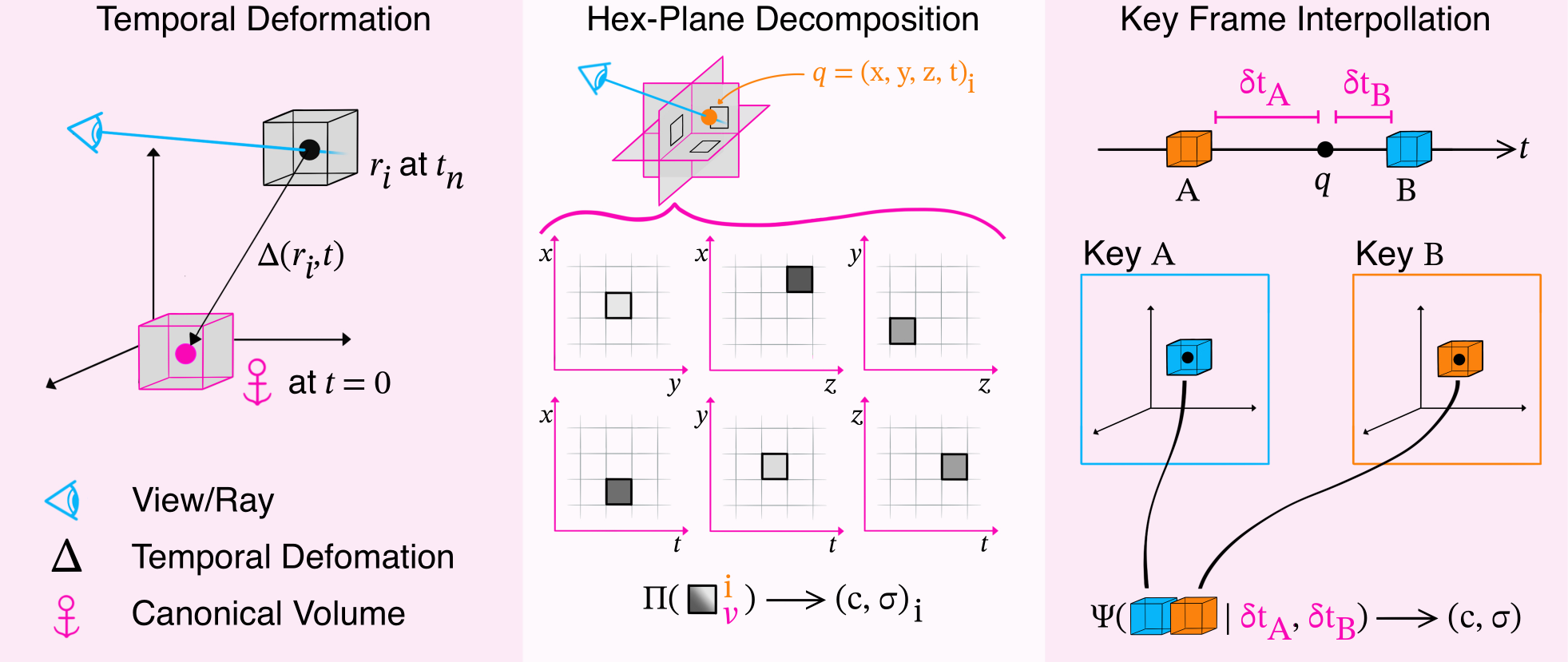}
    \Description{Three columns are used to identify the three main approaches to dynamic NVS. The first column shows the temporal deformation approach where an eye looks at a cube-volume and the volume's displacement at \begin{math}t=0\end{math} is illustrated as a 3D translation. The second column shows the hexplane decomposition where an eye looks at a point-volume and the six plane features pertaining to the point are used to determine color and density. The third column shows key-frame interpolation where two neighboring key frames along a time-line are combined to estimate the colour and opacity of a volume that in exists \begin{math}\delta_A\end{math} units of time after key frame A and \begin{math}\delta_B\end{math} units of time before keyframe B.}
    \caption{Illustrations of the \textbf{temporal deformation}, \textbf{hex-plane decomposition} and \textbf{key-frame interpolation} approaches for dynamic neural 4D representation}
    \label{fig: d-representations}
\end{figure*}

To begin, we can model a ``canonical'' (static) field with \begin{math}\Theta(\cdot)\end{math}, anchored at \begin{math}t=0\end{math}, and use a ``deformation'' field, 
\begin{equation}
    \Delta(\mathbf{r_i}, t) \rightarrow (\delta x, \delta y, \delta z),
\end{equation}
to model volumetric motion \cite{Pumarola_Corona_Pons-Moll_Moreno-Noguer_2021, Li_Slavcheva_Zollhoefer_Green_Lassner_Kim_Schmidt_Lovegrove_Goesele_Newcombe_2022, Yang_Gao_Zhou_Jiao_Zhang_Jin_2023}, whereby volumes are translated in time using \begin{math}r_i' = r_i + \Delta(\cdot)\end{math}; as illustrated in Figure \ref{fig: d-representations}. The deformation field is handled by an MLP \cite{Pumarola_Corona_Pons-Moll_Moreno-Noguer_2021, Li_Slavcheva_Zollhoefer_Green_Lassner_Kim_Schmidt_Lovegrove_Goesele_Newcombe_2022} meaning that motion is continuous in time. Therefore, as long as the training data captures smooth motions, slow-motion shots are easy to produce. Moreover, using an MLP reduces the model size significantly, making it efficient for compression and communication tasks. The drawbacks are long inference time and quality degradation for longer scenes. Additionally, this approach has trouble adapting to temporal changes in scene topology, so would not be suitable for filming objects that are not present during the anchor point.

Alternatively, we can decompose the 4D scene into discretized low-rank components \cite{Cao_Johnson_2023, Fridovich-Keil_Meanti_Warburg_Recht_Kanazawa_2023, Wu_Yi_Fang_Xie_Zhang_Wei_Liu_Tian_Wang_2023, Xu_Peng_Lin_He_Sun_Shen_Bao_Zhou_2023, Duisterhof_Mandi_Yao_Liu_Shou_Song_Ichnowski_2023, Huang_Sun_Yang_Lyu_Cao_Qi_2024}, e.g. grid-planes \begin{math}\mathbf{v} = \{xy, xz, yz, xt, yt, zt\}\end{math}. Features sampled from each plane, \begin{math}f_{\mathbf{v}}(\mathbf{q})\end{math}, for the point \begin{math}\mathbf{q} = (x,y,z,t)\end{math} are fused and decoded into color and density values, illustrated by \begin{math}\prod(\cdot)\end{math} in Figure \ref{fig: d-representations}. This is not dependent on a temporal anchor, so temporal changes in topology are not a problem. \citet{Fridovich-Keil_Meanti_Warburg_Recht_Kanazawa_2023} also show that the hex-plane decomposition can be adapted to work with datasets captured using various camera settings. However, as this approach discretizes the spatial axis, the scene is bounded by scale, so large or outdoor scenes are not feasible. Discretizing the temporal axis also results in temporal-jitter effects in the presence of fast motions, so may not be as effective as the prior approach for producing slow-motion shots.

Another option is to interpolate key-frame static representations with respect to time \cite{Song_Chen_Li_Chen_Chen_Yuan_Xu_Geiger_2023, Attal_Huang_Richardt_Zollhoefer_Kopf_OToole_Kim_2023, Wang_Hu_He_Wang_Yu_Tuytelaars_Xu_Wu_2023}. This models a different static field at each key-frame and interpolates between neighboring frames to produce continuous motion, as shown in Figure \ref{fig: d-representations}. Thus, we avoid temporal-jitter and handle geometric and visual changes in time. There is also a lower likelihood of quality degradation with longer scenes, and if we select a compact static key-frame representation data transmission (e.g. live broadcasting) can be facilitated \cite{Wang_Hu_He_Wang_Yu_Tuytelaars_Xu_Wu_2023}. However, this results in larger model size and requires significantly more computation and time to train.

Grouping the remainder of existing work, the last option is to directly parameterize a 3D representation in time \cite{Gan_Xu_Huang_Chen_Yokoya_2023, Fang_Yi_Wang_Xie_Zhang_Liu_Nießner_Tian_2022, Luiten_Kopanas_Leibe_Ramanan_2023, Sun_Jiao_Li_Zhang_Zhao_Xing_2024, Bae_Kim_Yun_Lee_Bang_Uh_2024, Yang_Yang_Pan_Zhang_2024}. For instance, \citet{Gan_Xu_Huang_Chen_Yokoya_2023} employs a voxel grid and parameterizes each voxel-feature in time to produce changes in texture and density. Comparatively, \citet{Yang_Yang_Pan_Zhang_2024} adapt the GS rendering function for 4D by proposing new 4D Gaussian primitives that directly optimize position, rotation and appearance without requiring spacetime decomposition, as seen with the hex-plane or canonical and deformation decompositions.

\subsection{Dynamic Scenes vs. Articulated 3D Objects}\label{sec: dynamic and animatable human pose}
A cinematographer may focus on capturing the entire dynamic scene or specifically on an articulated target, like a human or an animal. Capturing an entire scene increases complexity as it requires learning both foreground and background features consistently. Instead, focusing on object-centric reconstruction may simplify the problem if our only objective is to produce a rigged asset (i.e. an articulated object).

\subsubsection{Dynamic Scenes} Synthesizing entire scenes relies on good foreground and background color and density predictions. Here, the main challenges are dealing with dynamic backgrounds and unbounded spaces (e.g. outdoor).

Bounded scenes are typically contained inside a unit sphere or axis-aligned bounding box. As they are easier to reconstruct than unbounded scenes, more methods have been developed with bounded scenes: for NeRF \cite{Cao_Johnson_2023, Fridovich-Keil_Meanti_Warburg_Recht_Kanazawa_2023, Shao_Zheng_Tu_Liu_Zhang_Liu_2023,Gan_Xu_Huang_Chen_Yokoya_2023, Wang_Hu_He_Wang_Yu_Tuytelaars_Xu_Wu_2023, Fang_Yi_Wang_Xie_Zhang_Liu_Nießner_Tian_2022}, and for GS \cite{Sun_Jiao_Li_Zhang_Zhao_Xing_2024, Wu_Yi_Fang_Xie_Zhang_Wei_Liu_Tian_Wang_2023, Duisterhof_Mandi_Yao_Liu_Shou_Song_Ichnowski_2023, Huang_Sun_Yang_Lyu_Cao_Qi_2024, Yang_Yang_Pan_Zhang_2024}. These models are usually explicit representations as they produce higher quality results and are faster to train. Still, they have trouble modeling the out-of-bounds scene. Implicit dynamic NeRFs \cite{Pumarola_Corona_Pons-Moll_Moreno-Noguer_2021, Li_Slavcheva_Zollhoefer_Green_Lassner_Kim_Schmidt_Lovegrove_Goesele_Newcombe_2022} and GS \cite{Yang_Gao_Zhou_Jiao_Zhang_Jin_2023, Luiten_Kopanas_Leibe_Ramanan_2023, Bae_Kim_Yun_Lee_Bang_Uh_2024} are less constrained by a scene's bounds as they are continuous in space and time. However, these models are more likely to produce degenerate backgrounds in the presence of sparse views, meaning robust background prediction is still a challenge.

Unfortunately, there are currently no approaches that simultaneously tackle dynamic foregrounds and backgrounds, so cinematographers could choose to simplify the problem by assuming the background to be static \cite{Luiten_Kopanas_Leibe_Ramanan_2023}. Practical solutions are also possible, e.g. \citet{Huang_Sun_Yang_Lyu_Cao_Qi_2024} propose pre-generating dynamic masks using MiVOS \cite{cheng2021mivos} which enables their model (SC-GS) to separate the learning dynamic and static features - enhancing the visual quality of all dynamic elements.


\subsubsection{Articulated Humans/Objects} 
We can simplify the dynamic NVS paradigm by decomposing a scene into a canonical 3D human/object model and a pose generator. However, this relies on the pre-generation of an image-mask, for removing the background from a training dataset, so quality may depend on the robustness of the masking strategy. 

For those using rigged assets with conventional modeling software, mesh-based neural solutions \cite{Yang_Sun_Jampani_Vlasic_Cole_Chang_Ramanan_Freeman_Liu_2021, Yang_Sun_Jampani_Vlasic_Liu_Ramanan_Cole, Yang_Wang_Reddy_Ramanan_2023, Gao_Wang_Liu_Liu_Theobalt_Chen_2023, Yang_Vo_Neverova_Ramanan_Vedaldi_Joo_2022, He_Xu_Saito_Soatto_Tung_2021, Huang_Xu_Lassner_Li_Tung_2020, Zheng_Yu_Wei_Dai_Liu_2019} are a good option. Many methods use the skinned multi-person linear (SMPL) mesh model \cite{Loper_Mahmood_Romero_Pons-Moll_Black_2023} with a joint/skeletal model to control the pose, meaning meshes and animation rigs can be easily exported into commonly accepted formats (e.g. FBX). Nevertheless, mesh-based solutions produce worse reconstructions for non-rigid bodies, such as fuzzy hair, muscles flexing, etc.

Conversely, solutions using NeRF \cite{Xu_Alldieck_Sminchisescu_2021,Chao_Leung_2022,Su_Yu_Zollhoefer_Rhodin_2021,Gao_Yang_Kim_Peng_Liu_Tong_2022, Kwon_Kim_Ceylan_Fuchs_2021} or GS \cite{Zielonka_Bagautdinov_Saito_Zollhöfer_Thies_Romero_2023, Qian_Wang_Mihajlovic_Geiger_Tang_2024, Svitov_Morerio_Agapito_Del_Bue_2024, Wang_Zhang_Liu_Zhou_Zhang_Wu_Lin_2024, Kocabas_Chang_Gabriel_Tuzel_Ranjan_2023, Li_Tanke_Vo_Zollhöfer_Gall_Kanazawa_Lassner_2022} provide higher rendering quality and are better at modeling dynamic textures despite being inferior at handling rigid motions. To overcome this,  \citet{Huang_Sun_Yang_Lyu_Cao_Qi_2024} and \citet{ Yu_Julin_Milacski_Niinuma_Jeni_2024} propose learning a shallow hierarchy of control points for GS object representations. Furthermore, \citet{Wang_Zhang_Liu_Zhou_Zhang_Wu_Lin_2024} proposes a deeper hierarchy of control points reliant on semantic links between control points, e.g. linking a knee-joint to an ankle to facilitate a walking pose.

Past approaches to human reconstruction, \citet{Yang_Wang_Reddy_Ramanan_2023} propose category specific skeletal models for various mammals. Similarly, \citet{Li_Tanke_Vo_Zollhöfer_Gall_Kanazawa_Lassner_2022} propose an approach which does not rely on skeletal templates, making it applicable to other animals and joint-based objects. \citet{Loper_Mahmood_Romero_Pons-Moll_Black_2023,Chao_Leung_2022} and \citet{Gao_Yang_Kim_Peng_Liu_Tong_2022} also propose solutions to modeling multiple articulated people per-scene, as the general set of methods still assume single-person targets.


\subsection{Data Acquisition}\label{sec:data acquisition}
Data acquisition is key for generating a high-quality dynamic scene; despite this it is rarely discussed in NVS literature. In contrast to static NVS, where single-view camera (SVC) and multi-view camera (MVC) capture configurations are comparable, SVC and MVC datasets differ for dynamic scenes. MVC produces multiple observations per time step, while SVC is limited to single observations. Despite the advantage of MVC, more cameras incur higher costs and are difficult to move around during capture. 

Overcoming this requires application-specific solutions. For example, a popular way of reducing complexity for MVC datasets is to use a forward facing (FF) camera formation - where cameras are distributed in the same region of space and point in the same direction. FF scenes limit NVS to \begin{math}< 180\end{math} degrees of viewing freedom meaning shots that require 6 degrees of freedom (DoF), e.g. the fly-through shot in Figure \ref{fig: fly-through}, would not be possible. Instead, FF scenes may be a practical for tasks such as re-colorization, defocusing/re-focusing and generating smooth camera trajectories (within the limited capture space).

A practical way of attaining 6 DoF with a MVS or sparse-view set-up is to use depth sensors to enable more accurate geometric observations, or to support geometry initialization, e.g. initializing a GS point-cloud. However, if depth information cannot be obtained directly, another option is to use neural solutions for approximating depth from different viewpoints \cite{Fan_Cong_Wen_Wang_Zhang_Ding_Xu_Ivanovic_Pavone_Pavlakos_etal_2024,Wang_Leroy_Cabon_Chidlovskii_Revaud_2023}. While these methods do not produce optimal ground truth for depth, they can be effective for supporting geometric initialization and are especially tailored to sparse-view datasets. COLMAP \cite{schoenberger2016sfm,schoenberger2016mvs} has become the standard for calibrating cameras and generating initial points clouds for static and dynamic scenes. However, since this method is purposed for static scenes, there is a high likely hood of pose misalignment when most views consist of moving objects.

\section{Exhibition: \textit{An emotional sip of tea}}\label{sec:exhibit}
\begin{figure*}[ht]
    \centering
    \includegraphics[width=\linewidth]{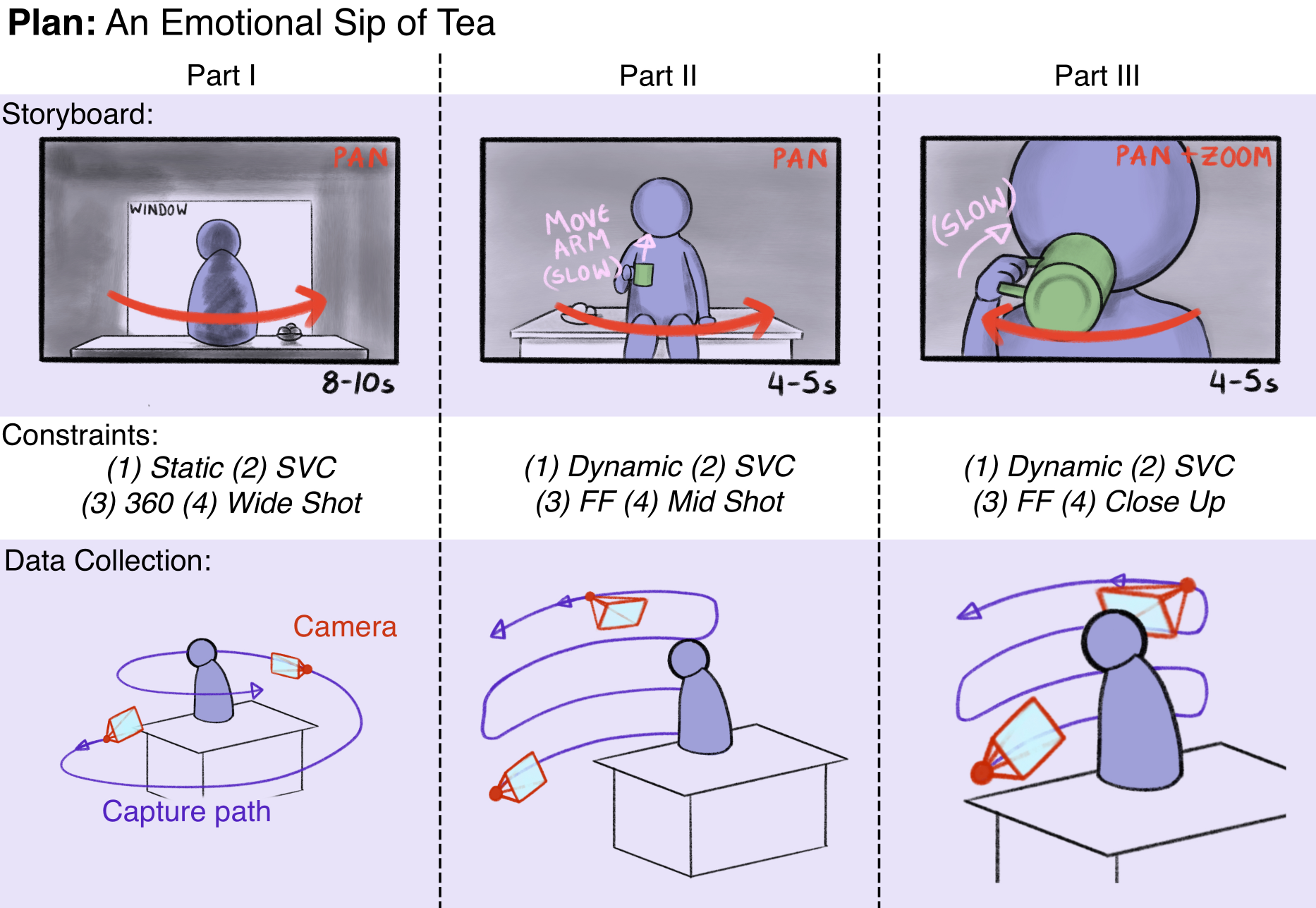}
    \Description{Three columns and three rows are shown. Each column represents Part 1, 2 or 3 of the video. The first row contains a sketch of the desired shot and the camera actions for each part. The second row contains the data capture constrains, where for Part 1 these are static, SVC, 360 and wide shot. For Part 2 and 3 these are Dynamic, SVC, FF and Mid shot/close up. The third row is an instruction sketch for data collection. For Part 1 the sketch shows a camera circulating around the target person. For Part 2 and 3 the sketch shows a camera moving around the front-side of the person at different distances from the person: for Part 2 this is far from the person, for Part 3 this is close up to the person.}
    \caption{\textbf{Production plan:} Our plan uses a storyboard for the actor and post-production staff to follow. We attached the scene constraints and visual plans for the camera man to collect each dataset}
    \label{fig: plan}
\end{figure*}
\begin{figure*}[t]
    \centering
    \includegraphics[width=\linewidth]{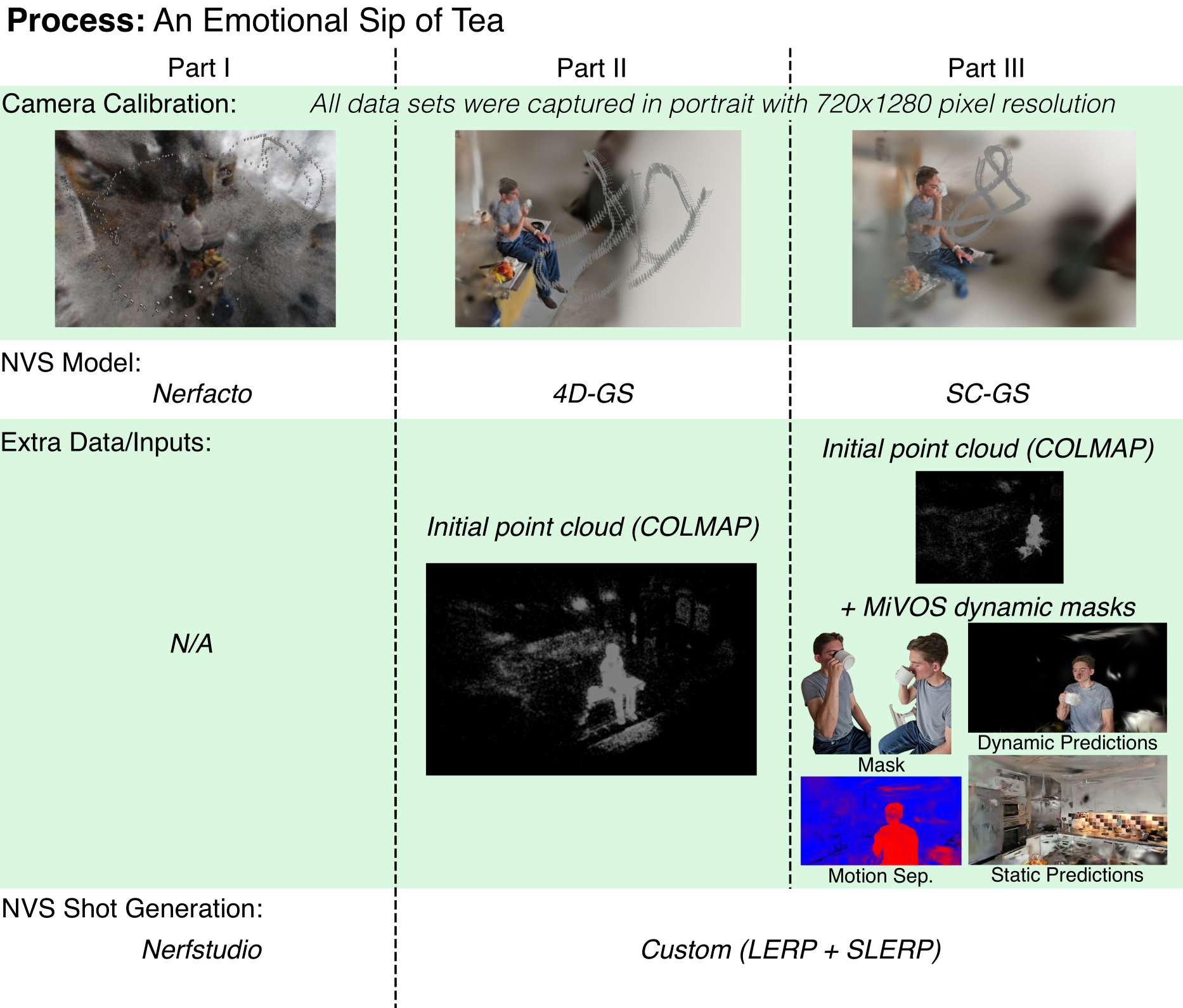}
    \Description{Three columns (for each part) and four rows are shown. Row 1 shows the distribution of the training cameras inside the 3D generated scenes. Row 2 identifies the NeRF or GS model used for generating each part. Row 3 identifies the extra inputs: for Part 1 this is empty, for Part 2 a point cloud is used, for Part 3 a point cloud and MIVOS dynamic masks are used. Row 4 idenitifies the methods used for shot generation. For Part 1 Nerfstudio NVS tools are used and for Part 2 and 3 we used a custom rendering process that involves LERP and SLERP.}
    \caption{\textbf{Production process:} Camera calibration, model selection, synthesizing additional inputs and generating novel shot trajectories. \textit{Motion Sep.:} we also show the results of separating dynamic (in red) and static (in blue) motion using the dynamic mask.}
    \label{fig: process}
\end{figure*}

To explore the capabilities of dynamic NVS in cinematography, we set out an example scene ``\textit{An emotional sip of tea}''  in this section. Like any cinematic production, we began with an objective: to depict an emotional young man drinking his tea. To achieve this, we broke the scene into three parts: (I) introducing the man in an emotional setting, (II) introducing the cup of tea, and (III) drinking the tea. These shots serve to illustrate various cinematographic needs, allowing us to exhibit three NVS models, highlighting their advantages and drawbacks. Our plan, process and results are provided in Figures \ref{fig: plan}, \ref{fig: process} and \ref{fig: results}, respectively.

Data acquisition was conducted using SVC with a mobile phone\footnote{A Samsung S23 was used}.
Part I involves a static 360-degree scene, which presents minimal challenges when using SVC. Complexity increases in Part II and III with dynamic scenes so we opt for FF shots to capture denser information with SVC. We subsequently derived the following shot schedule, also shown in Figure \ref{fig: plan}.
\begin{enumerate}
    \item Wide Shot + Pan (right) \begin{math}\rightarrow\end{math} Cut to Part II
    \item Mid Shot + Pan (left) \begin{math}\rightarrow\end{math} Cut to Part III
    \item Close-up + Pan (right) + Zoom \begin{math}\rightarrow\end{math} End
\end{enumerate}

\subsection{Part I} 
We selected the Nerfacto model provided by Nerfstudio \cite{nerfstudio} as it handles casually captured static scenes well. The Nerfstudio interactive viewer also provides numerous NVS tools, which we used to smooth the camera path and prolong the duration of Part I. The results show mostly show high-quality view-consistent details though there are some noisy results in regions that are sparsely viewed.

\subsection{Part II}
We selected 4D-GS \cite{Wu_Yi_Fang_Xie_Zhang_Wei_Liu_Tian_Wang_2023} as the scene is indoor (bounded) and captured with a FF formation. We chose this over NeRF alternatives as we were able to produce an initial point cloud using COLMAP. The results show the 4D-GS model is capable of handling view and time dependent lighting, e.g. on the actor's forearm. However, we also notice temporal jitter (more apparent in playback), which results from camera calibration errors from COLMAP.

\subsection{Part III}
We selected the SC-GS model as it has associated tools for pre-generating dynamic masks using MiVOS \cite{cheng2021mivos} and point clouds using COLMAP. Additionally, to mitigate the temporal jitter effects found in Part II, we utilized the SC-GS point-based editing tool to smooth the motions of various dynamic regions. The results show significantly less jitter and produce good results in sparse view regions, e.g. the left-side of the actor's face is only visible in \begin{math}15\%\end{math} of training dataset but still produces decent visual results. 

As the 4D-GS and SC-GS implementations do not contain NVS tools for generating and smoothing camera trajectories, we implemented our own method for rendering novel shots using linear interpolation (LERP) for interpolating between view position and spherical LERP (SLERP) for interpolating view rotation.

\begin{figure*}[t]
    \centering
    \includegraphics[width=\linewidth]{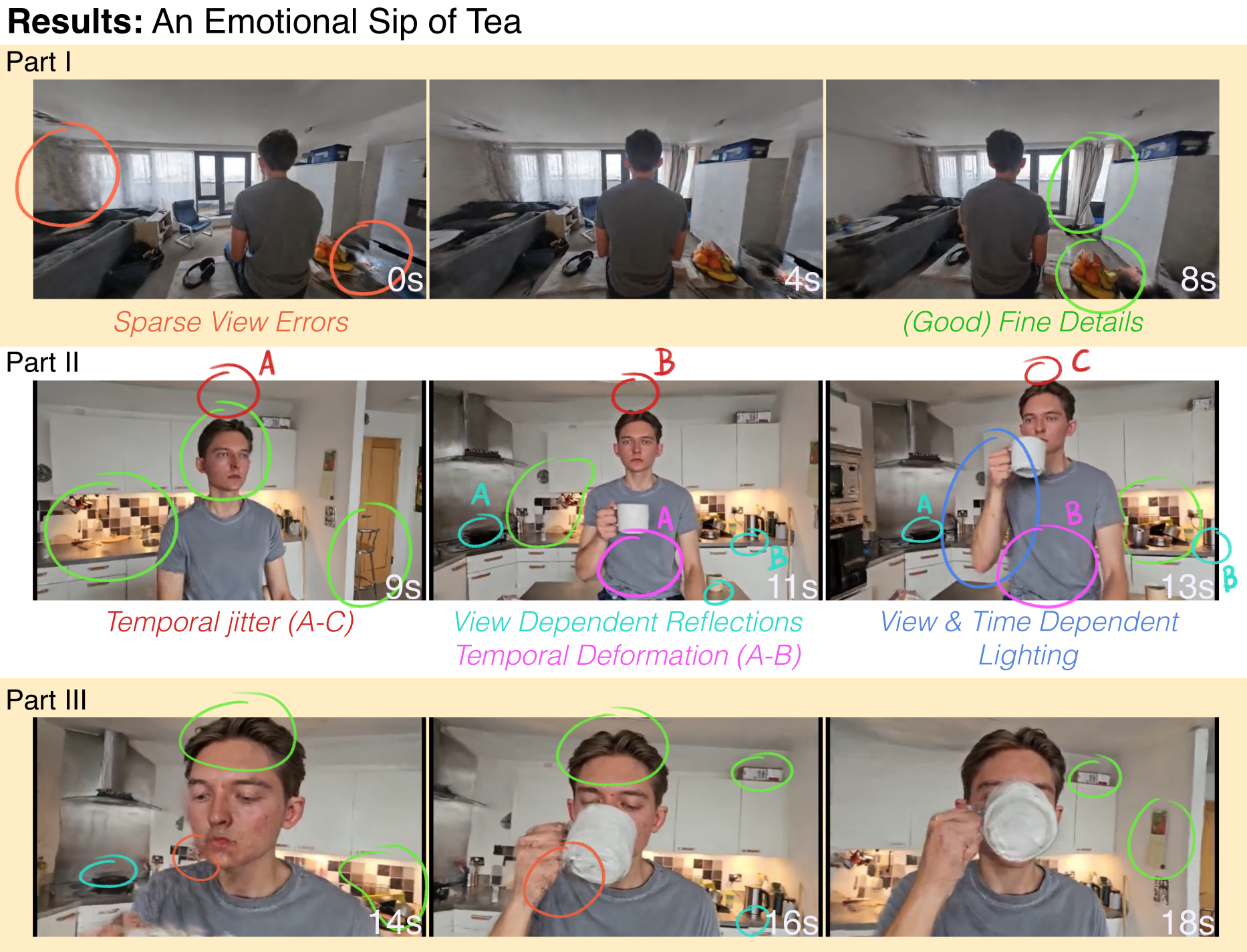}
    \Description{The rendered image results for each part are shown and split into three rows respective of each part. For part 1, good fine details and sparse view errors (i.e. 3-D noise) are highlighted. For part 2, good fine details, temporal jitter, view-dependent reflections, temporal deformations and view and light dependent lighting is highlighted. For part 3, good fine details, reflections and sparse view errors are highlighted.}
    \caption{\textbf{Results:} We generated shots and cut together renders from Part I-III to produce an short filmic masterpiece. Circles with different colors show achievement of rendering: fine details (in green), temporal deformations (in purple), reflections (in cyan) and view and time-dependent lighting (in blue). Some areas need more refinement: sparse view errors (in orange), and temporal jitter (in red)}
    \label{fig: results}
\end{figure*}

\section{Discussion and Conclusion}
Our showcase demonstrates impressive results for a casually captured dataset that consists of \begin{math}< 900\end{math} images. For instance, the right arm contracting in Part II, Figure~\ref{fig: results}, demonstrates the ability to model non-rigid deformations. The quality of the background in our dynamic scenes is also high-quality and contains various view-dependent lighting effects despite the limited number of views. Both factors would not have been possible with classical photogrammetric (e.g. mesh-based) tools. While there remain several challenges, such as pose-misalignment for calibrating dynamic datasets or inferior quality for large and/or fast motions, we are confident that NVS dynamic solutions offer significant potential for enhancing cinematographic applications.

\bibliographystyle{ACM-Reference-Format}
\bibliography{sample-base}

\end{document}